\pdfoutput=1 
\documentclass[11pt, a4paper]{article}

\usepackage[utf8]{inputenc}
\usepackage{geometry}
\usepackage{amssymb}
\usepackage[T1]{fontenc}

\usepackage{parskip} 
\usepackage{setspace}
\usepackage{graphicx}
\usepackage{booktabs}
\usepackage{hyperref}
\usepackage{amsmath}
\usepackage{float}
\usepackage{longtable}
\usepackage{caption}
\usepackage{cite}
\usepackage{tabularx} 
\usepackage{tcolorbox} 
\usepackage{url}
\usepackage{authblk}
\usepackage{listings}

\usepackage{xcolor}

\title{\textbf{Blu-WERP (Web Extraction and Refinement Pipeline)} \\ \vspace{0.3em} \textbf{A Scalable Pipeline for Preprocessing Large Language Model Datasets}}

\author{
  \textbf{Gowtham} \quad \textbf{Sai Rupesh} \quad \textbf{Sanjay Kumar} \quad \textbf{Saravanan} \quad \textbf{Venkata Chaithanya} \\
  \vspace{0.5em}
  \texttt{BluBridge Research} \\
  \vspace{0.5em}
  \texttt{contact@blubridge.ai}
}
\date{}

\begin{document}

\maketitle

\begin{abstract}
\noindent High-quality training data is fundamental to large language model (LLM) performance, yet existing preprocessing pipelines often struggle to effectively remove noise and unstructured content from web-scale corpora. This paper presents Blu-WERP, a novel data preprocessing pipeline designed to optimize the quality of Common Crawl WARC files for LLM training. We demonstrate that Blu-WERP significantly outperforms established baselines including DCLM across multiple model scales and evaluation benchmarks. Our pipeline processes Common Crawl WARC dumps, implementing advanced filtering and quality assessment mechanisms. We conducted comprehensive evaluations using models with 150M, 400M, 530M, 750M, and 1B parameters, testing against nine standard benchmarks categorized as World Knowledge \& Reasoning (MMLU, ARC-Easy, ARC-Challenge), Language Understanding (HellaSwag, Winogrande), and Commonsense Reasoning (PIQA, SocialIQA, CSQA, OpenBookQA). Results show Blu-WERP consistently achieved superior performance across all model scales. At the 1B parameter scale, Relatively Blu-WERP demonstrates a 4.0\% and 9.5\% aggregate improvement over DCLM and Fineweb respectively, while achieving quality-per-token efficiency gain. Categorical analysis reveals 2.4\% improvement in World Knowledge \& Reasoning, 6.2\% improvement in Language Understanding, and 4.2\% improvement in Commonsense Reasoning. These results establish Blu-WERP as a state-of-the-art preprocessing pipeline that substantially improves LLM training data quality and downstream model performance with reduced computational cost. Our findings contribute to the growing body of research on data-centric AI, demonstrating that preprocessing pipeline design significantly impacts LLM capabilities. The Blu-WERP pipeline represents a practical advancement in data quality optimization, offering researchers and practitioners an effective solution for improving LLM training efficiency and model performance.
\end{abstract}

\section{Introduction}
Large language models (LLMs) increasingly depend on the quality and structure of their available training data rather than architectural innovation \cite{56, 34}. While recent advances \cite{1, 2, 3, 4} have improved data filtering and deduplication, most existing pipelines still struggle to maintain a consistent balance between large-scale cleaning, semantic relevance, and downstream model performance. Many available corpora either over-filter informative content or retain excessive noise, leading to inefficiencies in training and limited generalization.

Our objective is to design a comprehensive data processing pipeline that ensures every stage from raw web data extraction to final corpus curation, produces training data that is clean, diverse, and semantically rich. The central goal is to demonstrate that models trained on our curated corpus achieve higher accuracy and robustness than those trained on widely-available datasets with less computational power. To achieve this, we integrate rigorous filtering, multi-level deduplication \cite{40}, and classifier-based quality control into a unified, scalable pipeline. The pipeline’s effectiveness is validated through controlled ablation studies and benchmark-driven evaluations, showing measurable gains in both dataset quality and model performance \cite{2, 50}. This work highlights that systematic data curation, rather than scale alone, is key to advancing language model capability.

While several filtering and deduplication pipelines have been proposed, such as C4 \cite{4}, OSCAR \cite{51}, and RefinedWeb \cite{3}, most focus on local quality metrics such as readability scores, token length, or repetition ratios. These measures capture only surface-level text quality and fail to reflect the global downstream impact of data on model generalization and benchmark performance. Furthermore, these pipelines often rely on static heuristic rules that lack adaptability across diverse web domains, languages, and data distributions.

To address these challenges, our pipeline incorporates semantic-aware and benchmark-driven filtering stages powered by a FastText \cite{10}-based classifier and the BETR (Benchmark Targeted Ranking) method \cite{57}. This combination enables dynamic, feedback-oriented filtering that prioritizes data with high semantic richness and proven benchmark alignment, resulting in a corpus that scales effectively while preserving performance consistency.

In addition to empirical evaluations, we incorporate a scaling-law-based dataset selection protocol \cite{34, 56} to assess how our curated corpus behaves at larger model capacities. Rather than relying solely on 150M–1B model results, we employ both a single-scale baseline and a two-step multi-scale extrapolation that models compute–loss and loss–accuracy trends across a size ladder \cite{49}. This enables prediction of dataset performance at target scales without training full-size models. The analysis shows that our dataset follows a more favorable scaling trajectory than DCLM \cite{2}, reinforcing its advantages even when projected to multi-billion-parameter LLMs (see Appendix \ref{app:scaling}).

\section{Background}
This work examines the problem of assembling large-scale text datasets suitable for training autoregressive Transformer-based \cite{28} language models. Such models learn to predict the next token in a sequence from the preceding context, a simple objective that enables broad adaptability across many natural language tasks once training is complete \cite{33}. The effectiveness of these models is strongly influenced by the data used during the initial pretraining stage, where they are exposed to vast amounts of unstructured text \cite{56, 34}. Later stages, such as task-specific fine-tuning or alignment through human feedback, refine specific capabilities, but the general linguistic competence of the model is determined primarily during pretraining \cite{4, 33}. Our focus is therefore on how to construct pretraining corpora that are both extensive and reliable, emphasizing approaches that promote data diversity and minimize noise.

In our setting, the raw text originates from Common Crawl WARC records \cite{5}, which provide large-scale snapshots of publicly available web content. Although this source offers substantial breadth and diversity, the data in its raw form is noisy, inconsistent, and unsuitable for direct use in pretraining. Web pages often contain boilerplate, navigation text, advertisements, duplicated material, and content of highly variable quality - issues extensively documented in prior corpus construction efforts \cite{6, 8, 9}. Because of this heterogeneity, the effectiveness of any large-scale language model depends not only on the volume of text collected, but also on the quality of the processing pipeline that converts raw WARC records into clean, structured training data \cite{4, 36}. Our work focuses on designing and evaluating such a pipeline, emphasizing systematic filtering, normalization, and data quality controls that ensure the resulting corpus is both scalable and reliable for pretraining.

While large-scale web archives contain ample raw text for pretraining, the effectiveness of that data depends heavily on how it is filtered and structured before use \cite{4, 36}. Web text includes substantial noise such as boilerplate HTML fragments, low-information chatter, templated pages, and other forms of unnatural language that do not reflect the contexts in which models are expected to perform \cite{4, 39}. Removing such noise is necessary, yet filtering must be done carefully; eliminating too aggressively risks discarding the very material that contributes to reasoning ability, knowledge coverage, and educational value \cite{39}. Duplication poses a similar challenge, as repeated content can appear at the document, paragraph, and line levels, and must be handled with multiple, coordinated deduplication strategies \cite{40}.

Our pipeline is designed to maintain this balance: it aims not only to clean the data, but to preserve and emphasize text that contributes meaningful knowledge and utility. To support this, we apply structured quality filtering followed by targeted repetition reduction, guided by a FastText-based classifier \cite{10, 57} that helps distinguish useful and informative content from low-value text, ultimately aligning the dataset with strong benchmark performance and real downstream usability.

Prior research on constructing pretraining corpora from web data has shown that pipeline design plays a critical role in determining final dataset quality \cite{4, 36, 39}. Different approaches vary in how they identify language, remove boilerplate text, filter for content quality, and reduce duplication \cite{6, 8, 36}. Some pipelines prioritize aggressive filtering to eliminate noise, while others aim to maximize dataset size by retaining more content \cite{4, 6}. Deduplication strategies also vary, ranging from exact matching at the document level to fuzzy, multi-granular approaches that address repeated patterns across pages and paragraphs \cite{40}. These variations highlight that there is no single universally optimal pipeline; instead, each design reflects trade-offs among cleanliness, dataset scale, content diversity, and practical compute constraints. Our work contributes to this line of research by emphasizing both high-quality filtering and the preservation of informative, educationally valuable content, supported by multi-level deduplication and classifier-driven selection to better balance data quality with data utility.

\section{Related Work}

\subsection{Web-scale pretraining corpora}
Early efforts established foundational practices for large-scale corpus construction. C4 \cite{4} introduced heuristic filtering of Common Crawl \cite{5} for the T5 model, though subsequent analyses highlighted limitations in static rules and the unintended removal of valid content \cite{39, 40}. CCNet \cite{36} advanced the field through systematic fastText-based language identification and Wikipedia-perplexity filtering, establishing methodological patterns that influenced many subsequent datasets. More recent open-source efforts demonstrate that carefully filtered web data alone can match or exceed curated mixtures: RedPajama \cite{37} reproduced LLaMA’s data distribution across multiple Common Crawl snapshots; RefinedWeb \cite{3} achieved state-of-the-art results using trafilatura-based extraction and MinHash deduplication \cite{14}; and FineWeb \cite{1} introduced FineWeb-Edu, a Llama-3.3-70B-annotated \cite{61} subset showing strong gains on reasoning benchmarks including MMLU \cite{18}, ARC \cite{19}, and OpenBookQA \cite{42}. Nemotron-CC \cite{47} further expanded high-quality data collection by applying ensemble model–based quality stratification to multiple Common Crawl snapshots. Finally, Dolma \cite{38}, paired with the OLMo framework \cite{35}, aggregated diverse sources with an emphasis on transparency, open tooling, and comprehensive documentation.

\subsection{Benchmark-driven evaluation and filtering}
Most pipelines evaluate corpora post-hoc, creating a disconnect between filtering criteria and downstream impact. DataComp-LM (DCLM \cite{2}) provided the first large-scale testbed with standardized training recipes and comprehensive downstream evaluations across Common Crawl data, demonstrating that model-based filtering outperforms static heuristics. Their DCLM-Baseline achieved 64\% 5-shot MMLU accuracy using a 7B parameter model, representing a 6.6 point improvement over MAP-Neo while using 40\% less compute. This work established rigorous benchmarking as essential for comparing data curation strategies.

\subsection{Content extraction and quality filtering}
Custom extraction significantly improves quality over WET baselines. JusText \cite{6} pioneered stopword-density-based boilerplate removal and remains competitive in modern pipelines. Trafilatura \cite{44} provides metadata-aware extraction with strong robustness across diverse page types and has been widely adopted in RefinedWeb and similar pipelines. Resiliparse \cite{9} emphasizes speed and resilience to malformed HTML, serving as DCLM’s default extractor \cite{2} with order-of-magnitude performance advantages. Gopher \cite{39} established document-level heuristics for length, repetition, and symbol ratios that have become standard quality filters across modern pipelines. Both FineWeb \cite{1} and DCLM \cite{2} show that classifier-based filtering that targets educational or high-quality content yields substantial gains over purely heuristic approaches. Nemotron-CC \cite{47} further demonstrates that removing traditional heuristic filters from data pre-selected by model-based classifiers improves dataset yield without degrading downstream performance.

\subsection{Deduplication strategies}
Deduplication has become standard practice to mitigate redundancy, reduce memorization risks, and improve training efficiency \cite{40}. Modern pipelines combine exact document hashing with MinHash-based near-duplicate detection \cite{14} to capture both byte-identical and paraphrastic overlaps within dumps and across Common Crawl snapshots \cite{3, 38}. Bloom Filters \cite{13} provide memory-efficient probabilistic membership tests for streaming deduplication, enabling scalable document-level and n-gram-level checks. RefinedWeb \cite{3}, FineWeb \cite{1}, Dolma \cite{38}, and Nemotron-CC \cite{47} all document multi-stage deduplication and decontamination as core components of their pipelines, providing reference implementations that illustrate the trade-offs between memory usage, strictness, and recall at web scale.

\section{Work Flow}
While prior work has explored these components - ranging from extractor choice \cite{6, 9, 44} and repetition heuristics \cite{39} to MinHash-based deduplication \cite{14} and classifier-guided filtering \cite{1, 2} - most pipelines investigate them in isolation, leaving open questions about how these design choices interact and how they influence downstream generalization. Recent evidence from DataComp-LM \cite{2} and FineWeb \cite{1} suggests that extraction quality, deduplication granularity, and semantic quality signals can meaningfully shift benchmark performance, particularly for small and mid-scale models. Motivated by these observations, we undertake a systematic examination of these axes through controlled parser ablations, hierarchical deduplication analyses, and classifier-driven quality studies. This perspective lays the foundation for a unified, benchmark-conditioned workflow that we detail next, operationalizing these insights into a practical pretraining data pipeline.

In this section, we present the workflow of the Blu-WERP Pipeline, detailing its core components, their functional roles within the processing sequence, and the series of ablation studies conducted for each module. We further describe the working mechanism of individual components and how they interact to collectively enhance data quality and downstream model performance.

\subsection{Text Extraction}
Text extraction is the first step in converting raw HTML from Common Crawl \cite{5} into clean text suitable for filtering and deduplication. To evaluate its impact, we compare four extraction methods: Resiliparse \cite{9}, jusText \cite{6}, Trafilatura \cite{44}, and the default WET pre-extracted text from Common Crawl \cite{5}. One notable difference is that jusText applies stopword-based filtering during extraction, which implicitly removes pages lacking recognizable language cues. In contrast, Resiliparse and Trafilatura do not enforce this constraint, resulting in higher initial retention immediately after parsing. However, after full pipeline processing which includes quality filtering and deduplication the jusText-based pipeline retains a higher proportion of useful and well-structured text (see Appendix \ref{app:text_extraction} for retention details).

\begin{table}[H]
\centering
\caption{Parser ablation results showing aggregate benchmark performance across four text extraction methods.}
\label{tab:parser_ablation}
\begin{tabular}{cc} 
\toprule
\textbf{Parser} & \textbf{Aggregate Score (\%)} \\
\midrule
Resiliparse & 44.02 \\
\textbf{Justext} & \textbf{44.74} \\
Trafilatura & 44.29 \\
WET & 40.57 \\
\bottomrule
\end{tabular}
\end{table}

To ensure a fair comparison across extractors, we applied the URL filtering and FastText-based language identification stage \cite{36} after extraction in all configurations. This eliminates differences caused by the presence or absence of built-in stopword heuristics and isolates the impact of extraction quality itself.

\subsection{Deduplication}
To improve the information density of the dataset, we perform deduplication using a Bloom Filter \cite{13} based approach. Web data often contains repeated or mirrored documents, and retaining these duplicates reduces the effective diversity of the training corpus. By removing duplicates, the model is exposed to more unique content for the same number of tokens, leading to stronger downstream performance \cite{40}. For deduplication, we explored multiple techniques, including Bloom Filter \cite{13}, MinHash \cite{14}, and Suffix Array based methods \cite{15}. Several configurations of the Bloom Filter were also evaluated to identify the most effective setup for large-scale text processing. Based on extensive experimentation, we selected the combination of deduplication methods that offered the best balance between computational efficiency and deduplication accuracy, ensuring optimal performance for our filtered corpus (see Appendix \ref{app:deduplication}).

\begin{table}[H]
\centering
\caption{Deduplication method comparison showing aggregate benchmark performance across four approaches.}
\label{tab:dedup_ablation}
\begin{tabular}{cc} 
\toprule
\textbf{Method} & \textbf{Aggregate Score (\%)} \\
\midrule
Exact+Min-Hash & 48.20 \\
Exact+Sub String+Min Hash & 49.15 \\
Bloom Filter (Both) & 47.83 \\
\textbf{Bloom Filter (Old Both)} & \textbf{49.22} \\
\bottomrule
\end{tabular}
\end{table}

\subsection{Classifier} 
After structural and repetition-based filtering, a significant portion of web text still remains that is syntactically clean but semantically shallow. Such text often lacks informational depth, instructional value, or meaningful knowledge content. To address this, we apply a semantic quality classifier as the final stage of the pipeline. The classifier is designed to distinguish informative and educationally useful documents from text that is generic, promotional, or low-value despite being well-formed \cite{10, 57, 63, 64}. We use a FastText-based classifier \cite{10} due to its ability to operate efficiently at scale and its robustness to vocabulary variation in web data. FastText incorporates subword-level representations, allowing it to generalize effectively across diverse linguistic domains. This enables reliable large-scale sorting of web documents into high-value and low-value content classes without introducing significant computational overhead. The semantic classifier serves a complementary role to earlier filtering stages. While previous filters remove noise, boilerplate, and repetition, the classifier targets conceptual usefulness. This ensures that the final dataset is not only clean and diverse, but also rich in content that supports effective language model pretraining and downstream reasoning performance (refer to Appendix \ref{app:classifier}).

\begin{table}[H]
\centering
\caption{Classifier ablation results comparing aggregate benchmark performance across four quality filtering approaches.}
\label{tab:classifier_ablation}
\begin{tabular}{cc} 
\toprule
\textbf{Classifier} & \textbf{Aggregate Score (\%)} \\
\midrule
Openherms 2.5 + Eli5 & 51.37 \\
\textbf{BETR} & \textbf{53.8} \\
Llama Score + BERT & 51.28 \\
DEBERTA & 49.48 \\
\bottomrule
\end{tabular}
\end{table}

\section{Experimental Setup}

\subsection{Training Configuration}
To ensure a fair comparison between extraction pipelines, every downstream experiment was conducted using a single, fixed model and identical training regimen. We selected the OLMo family \cite{35} and performed all ablations and final runs with the OLMo-1B model. All training runs used a context window of 2048 tokens, the OLMo tokenizer, and a global (effective) batch size of 2M tokens. Following the Chinchilla scaling heuristic \cite{56}, we trained each model for 20× the model parameter count in tokens; for the 1B-parameter model this corresponds to 20 billion training tokens per run. In total we executed all our runs distributed across H200 GPUs.

\subsection{Evaluation Protocol}
\subsubsection{Few-shot Configuration} 

Following the standardized evaluation protocol established in prior work \cite{33}, our evaluation framework employs few-shot prompting to convey task intent to the model. All tasks are evaluated under a 5-shot setting, ensuring consistency with prior literature and comparability across diverse benchmarks \cite{50}. The number of in-context examples and their sampling strategy are determined by each dataset’s established evaluation configuration. For each benchmark, the evaluation framework automatically selects five representative in-context examples from the corresponding dataset to construct the prompt. These examples are balanced across the label space to avoid bias toward any single class and are chosen deterministically using a fixed random seed to ensure reproducibility. Restricting the number of examples also keeps computational overhead manageable while aligning with widely adopted evaluation standards. Two newline characters separate in-context examples. The total input length, including both prompt and completion, is capped at 2,048 tokens, and a \texttt{<bos>} token is injected only for model families that explicitly require it.

\subsubsection{Benchmark Standardization} 

We adopt a fully documented, public evaluation standard for multiple-choice LLM benchmarks and implement it within a single open evaluation harness to ensure strict reproducibility across scales and ablations \cite{50}. Each multiple-choice benchmark contains structured fields such as the question, answer options, and, in some cases, additional context. To maintain consistency across all benchmarks, our pipeline applies a unified instance formatting scheme when constructing prompts for the model. Each question is prefixed with “Question:” and suffixed with “Answer:” to clearly define the task without relying on verbose instructions. For tasks that deviate semantically from a standard question–answer structure, the evaluation framework adapts the prefix to align with the dataset’s intent (e.g., “Fill in the blank:” or “Choose the best continuation:”). To ensure tokenizer compatibility, answer options are formatted with a prefixed space before each label (e.g., “ A.”, “ B.”, etc.), guaranteeing that the model’s predicted answer token aligns exactly with the expected label token.

\subsubsection{Sampling Strategy}

For evaluation, the framework employs a standardized sampling procedure. When labels are publicly available, the test split is used; otherwise, the validation split is selected. Datasets containing more than 1,500 instances are uniformly downsampled to 1,000 examples to maintain computational efficiency while preserving representativeness. This controlled sampling ensures that performance differences arise from model behavior rather than variations in dataset volume or formatting \cite{50}.

\subsubsection{Evaluation Constraints and Controls}

\textbf{Subject-wise Aggregation:} For multi-subject benchmarks such as MMLU \cite{18}, we evaluate all 57 subjects and compute macro-averaged scores across subjects, following the dataset’s official evaluation protocol \cite{18}.

\textbf{Exemplar Decontamination:} To isolate data-curation effects and prevent potential contamination between the embedding/bin training utilities and evaluation prompts, the 5-shot demonstration pools are made strictly disjoint from any sample used in the embedding-based “bin” training pipeline. Candidate examples that appear in the bin or embedding pool (checked by both ID and content) are blacklisted before finalizing the demonstration set. This follows recent findings showing that contamination can inflate benchmark scores and obscure true ablation effects \cite{53, 54}.
\newpage
\textbf{Implementation Notes:} All the above evaluation rules are implemented within a single unified harness, encompassing task adapters, stop-sequence handling, reporting scripts, and fixed seeds. This eliminates incidental variation across runs and enables bitwise-identical re-execution of all evaluations \cite{50}.

\subsection{Benchmark Suite and Rationale}
We retain nine datasets whose sensitivity profiles complement each other along the axes of retrieval, reasoning, coverage, and robustness.
\begin{itemize}
    \item \textbf{ARC-Challenge \& ARC-Easy} \cite{19}: Difficulty-stratified science questions that stress multi-hop retrieval and compositional reasoning; gradients across the two splits cleanly separate the impact of corpus cleaning from raw scale effects.
    \item \textbf{CommonsenseQA} \cite{41}: Requires prior world knowledge beyond surface statistics; score deltas therefore track corpus diversity rather than sheer token count.
    \item \textbf{HellaSwag} \cite{20}: Adversarially filtered completion pairs that expose shallow temporal heuristics; gains indicate improved event-level modelling and effective de-duplication.
    \item \textbf{MMLU} \cite{18}: 57 subject areas act as a coarse coverage barometer; well-suited for head-to-head scale comparisons under identical training recipes.
    \item \textbf{OpenBookQA} \cite{42}: Explicit fact-plus-knowledge integration; upticks signal better linkage between surface facts and implicit knowledge captured during pre-training.
    \item \textbf{PIQA} \cite{22}: Physical commonsense and affordance reasoning; shifts reflect the inclusion of procedural / how-to style text in the pre-training mix.
    \item \textbf{Social-IQa} \cite{43}: Theory-of-mind inference (motives, reactions); provides a counterpart to PIQA and quantifies social-reasoning signals introduced by data choices.
    \item \textbf{WinoGrande} \cite{26}: Adversarial coreference resolution benchmark engineered to suppress lexical shortcuts; performance correlates with representation robustness rather than n-gram memories.
\end{itemize}

\section{Results}

\begin{figure}[H]
\centering
\includegraphics[width=\linewidth]{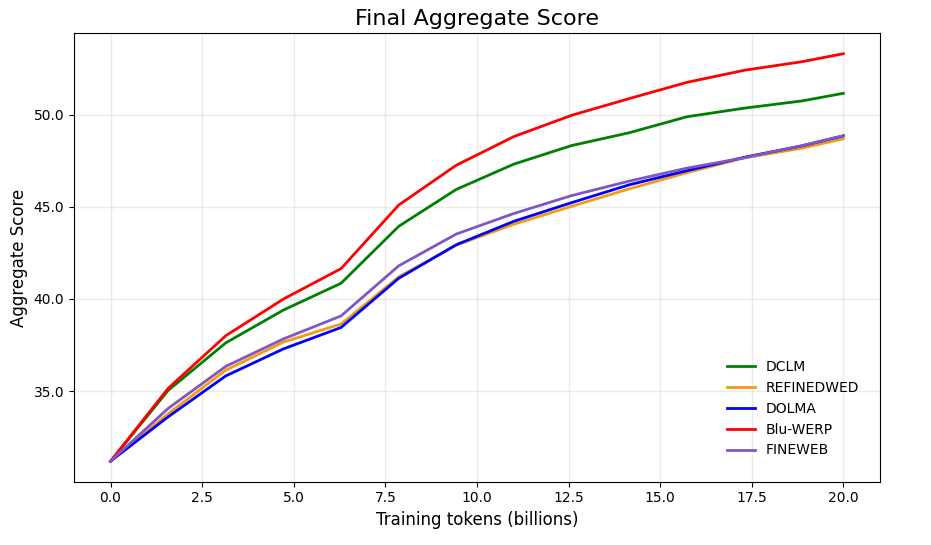}
\caption{\textbf{Aggregate benchmark performance comparison across five datasets. }Blu-WERP achieves 53.88\% aggregate accuracy, outperforming DCLM \cite{2} (51.81\%) and other baselines.}
\label{fig:aggregate_score}
\end{figure}

We evaluate our pipeline by comparing it with 4 other datasets. To make the comparison meaningful, we rely on held-out downstream curated dataset against widely used public pretraining corpora, including DCLM \cite{2}, Fineweb \cite{1} benchmark performance rather than dataset size or number of documents alone. This allows us to assess the quality of the retained content rather than simply its volume. The evaluation suite spans common language understanding and reasoning tasks, measuring both general capability and robustness to noisy input \cite{18, 19, 20, 22, 41, 43, 50}.

Across the aggregate benchmark score, our pipeline consistently surpasses prior open-source pipelines. Most notably, we achieved a score of 53.38 while DCLM \cite{2} scored just 51.15, which is currently regarded as a state-of-the-art public pipeline. The performance gains indicate that our filtering, repetition control \cite{40}, and classifier-based quality reinforcement \cite{57} collectively produce a more information-dense corpus. Importantly, these improvements are not merely due to aggressive pruning; the resulting dataset retains educational, factual, and diverse sources of knowledge, preserving the breadth required for general-purpose language model training.

\begin{figure}[H]
\centering
\includegraphics[width=\linewidth]{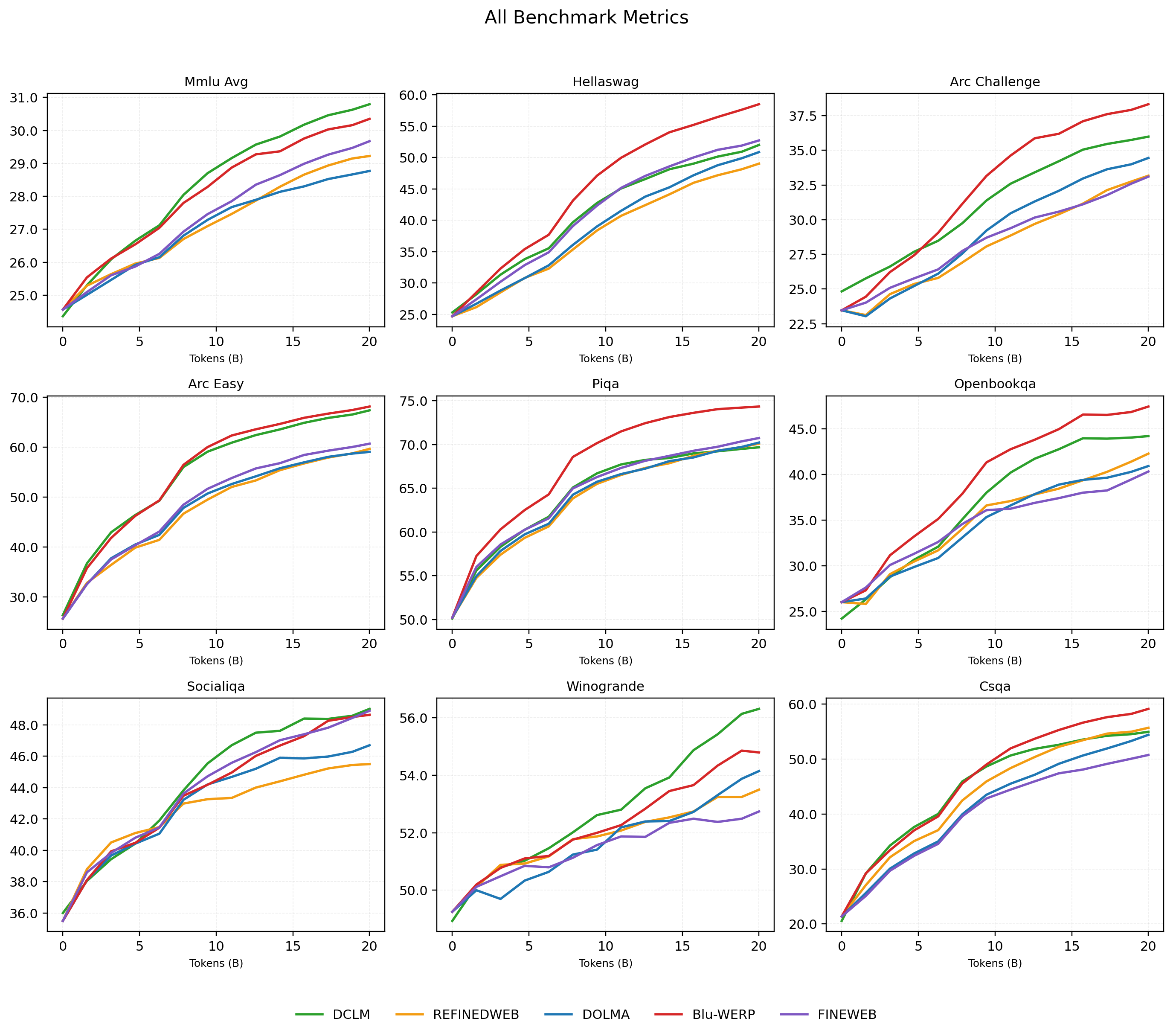}
\caption{\textbf{Evaluation results across nine benchmarks from the standardized evaluation suite.} Our dataset outperforms all other corpora on the majority of tasks, with competitive results comparable to DCLM. While performance on MMLU and SocialIQA slightly trails DCLM, our dataset achieves parity in benchmarks assessing world knowledge (MMLU, ARC Easy, ARC Challenge) and demonstrates superior results in language understanding (HellaSwag, SocialIQA) and common-sense reasoning (CSQA, PIQA).}
\label{fig:all_benchmarks}
\end{figure}

These results indicate that the proposed pipeline produces a dataset with balanced coverage and strong generalization across diverse reasoning and understanding tasks.
\newpage
\section{Conclusion}
We present Blu-WERP, a novel preprocessing pipeline that significantly improves the quality of Common Crawl data \cite{5} for language model pretraining. Through systematic ablations across extraction methods, deduplication strategies \cite{40}, and classifier-based filtering, we demonstrate that pipeline design choices substantially impact downstream model performance. At the 1B parameter scale, Blu-WERP relatively achieves a 4\% and 9.5\% aggregate improvement over state-of-the-art baselines (DCLM \cite{2} and FineWeb \cite{1}) respectively, while achieving quality-per-token efficiency gain, establishing a higher signal-to-noise ratio in the curated corpus. Our hierarchical Bloom filter \cite{13} deduplication reduces memory footprint while maintaining competitive performance, and the implementation of BETR-driven \cite{57} classifier filtering enables dynamic, benchmark-aligned document selection. These results validate that data-centric optimization - rather than scale alone is key to advancing language model capabilities. The Blu-WERP pipeline offers researchers a practical, reproducible framework for improving training efficiency and model quality, with all components publicly documented to support further exploration in data-centric AI.

\section{Limitations \& Future Work}
\textbf{Language and domain scope} 

Our evaluation focuses exclusively on English web data and general-purpose benchmarks. Extension to multilingual corpora and domain-specific evaluations (e.g., code, scientific literature) remains unexamined and may require adapted quality classifiers.

\textbf{Decontamination} 

We did not perform explicit decontamination of evaluation benchmarks from our training corpus, following the protocol of DCLM \cite{2} and FineWeb \cite{1}. While our 5-shot evaluation setup minimizes direct contamination risks, rigorous fuzzy-deduplication against benchmark data would strengthen confidence in reported gains \cite{53,54}.

\textbf{Factual accuracy} 

Our pipeline does not verify factual correctness of retained documents. While classifier-based filtering prioritizes educational content, it cannot guarantee truthfulness, potentially retaining misinformation present in the original web crawl.

\newpage
\nocite{*} 
\bibliographystyle{unsrt}
\bibliography{references}

\newpage

\appendix
\section*{APPENDIX}
\addcontentsline{toc}{section}{Appendix}

\section{Data Ablation Setup}

\begin{table}[H]
\centering
\caption{Model Architecture}
\label{tab:model_arch}
\begin{tabular}{ll}
\hline
\textbf{Parameter} & \textbf{Value} \\
\hline
Architecture & OLMo-1B \\
Number of attention heads & 16 \\
Number of hidden layers & 16 \\
Number of key-value heads & 16 \\
RMS Norm epsilon & 1.00E-06 \\
Tied word embeddings & False \\
Embedding size & 50,304 \\
Total number of parameters & 1.03B  \\
Random initialization std & 0.02 \\
Tokenizer & OLMo 7B \\
\hline
\end{tabular}
\end{table}

\begin{table}[H]
\centering
\caption{Distributed Training Setup}
\label{tab:dist_training}
\begin{tabular}{ll}
\hline
\textbf{Parameter} & \textbf{Value} \\
\hline
Data parallelism (dp) & 8 \\
Tensor parallelism (tp) & 1 \\
Pipeline parallelism (pp) & 1 \\
Micro-batch size & 16 \\
Sequence length & 2048 \\
Batch accumulation per replica & 6  \\
\hline
\end{tabular}
\end{table}

\begin{table}[H]
\centering
\caption{Optimizer Configuration}
\label{tab:optimizer}
\begin{tabular}{ll}
\hline
\textbf{Parameter} & \textbf{Value} \\
\hline
Adam beta1 & 0.9 \\
Adam beta2 & 0.95 \\
Adam epsilon & 1.00E-08 \\
Gradient clipping & 1 \\
Weight decay & 0.1 \\
Learning rate & 2.10E-03 \\
Warmup style & Linear warmup \\
Decay style & Cosine \\
\hline
\end{tabular}
\end{table}
\newpage
\section{Text Extraction} \label{app:text_extraction}
Web crawl archives are commonly distributed in two canonical formats: WARC \cite{5}, which contains raw HTTP responses including full HTML (tags, headers, footers and site chrome), and WET \cite{5}, which contains text extracted from WARC by a basic HTML pass.
WARC records preserve the original page structure and chrome (navigation bars, banners, page numbers, etc.), elements that are typically undesirable for pretraining.
Although WET reduces the degree of markup, it is generated by a simple extraction step and can still include substantial non-informative boilerplate.

\textbf{Parsers and experimental motivation} 

For the Parser Ablations, we evaluated multiple HTML parsing frameworks to assess their effectiveness in extracting clean textual content from web data.
Specifically, we experimented with Justext \cite{6}, Resiliparse \cite{9}, and Trafilatura \cite{44}, and compared their outputs against the WET baseline \cite{5}.
To ensure a fair and consistent comparison, we applied only language filter after parsing, isolating the impact of the parser itself on data quality.
Furthermore, retention rates were measured post-filtering in terms of tokens to analyze how each parser influenced the volume and quality of retained text within our pipeline.

\textbf{Language control and experimental protocol} 

Because some extractors apply implicit or explicit language heuristics (for example, jusText uses stopword-based paragraph scoring and, by default, can exclude non-English paragraphs \cite{6}), we applied an explicit language identification filter to all parser outputs to ensure a fair comparison.
Specifically, we used a FastText language identifier with a conservative confidence threshold (0.65) \cite{10} to normalize language composition across outputs.
After language filtering, identical downstream preprocessing (tokenization, deduplication, and sharding) and model training procedures were applied to each corpus variant.

\textbf{Evaluation methodology} 

For each parser output we measured (a) retention statistics (documents and tokens retained after filtering and deduplication) and (b) downstream performance on a fixed benchmark suite using the same model architecture and training hyperparameters.
This design isolates the contribution of the extraction stage from subsequent modeling and optimization choices.

\begin{table}[H]
\centering
\caption{Comparison of retention for different parsers}
\label{tab:parser_retention}
\begin{tabular}{lc}
\toprule
\textbf{Parser} & \textbf{Retention(After filters)} \\
\midrule
Resiliparse & 19.44\% \\
\textbf{Justext} & \textbf{49.96}\% \\
Trafilatura & 43.25\% \\
\bottomrule
\end{tabular}
\end{table}

We initially applied parsers and parsed the data, and processed through our pipeline till the filters to get the correct retention rate.

\begin{figure}[H]
\centering
\includegraphics[width=\linewidth]{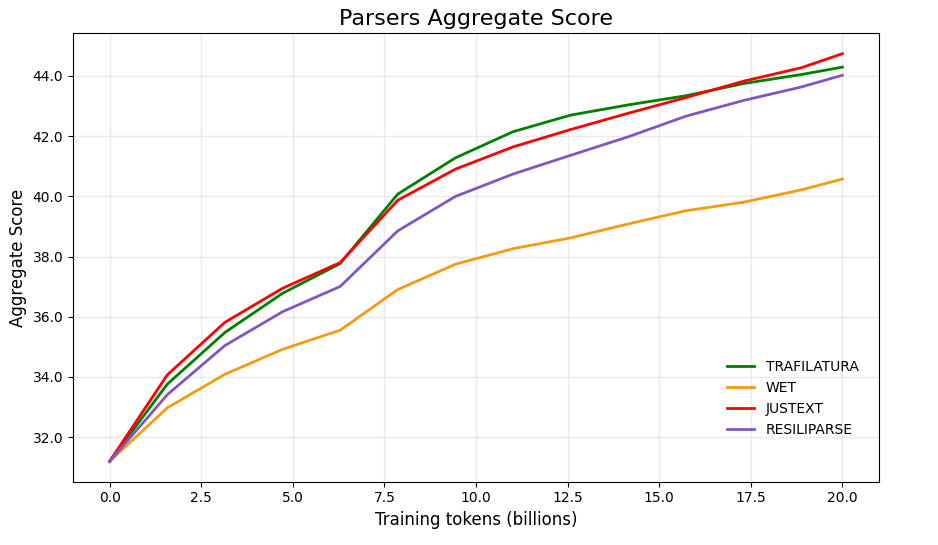}
\caption{\textbf{Aggregate benchmark performance across four parser configurations.} Justext achieves highest aggregate score (0.4474) with 49.96\% retention, followed by Trafilatura (0.4429, 43.25\% retention), Resiliparse (0.4402, 19.44\% retention), and WET baseline (0.4057).}
\label{fig:parser_score}
\end{figure}

\textbf{Results (summary)}

Across the parser ablations, Justext achieved the best overall results, with the highest aggregate score (0.4474) and a strong retention rate (49.96\%).
Trafilatura followed closely with an aggregate score of 0.4429 and 43.25\% retention, while Resiliparse scored 0.4402 with a lower retention of 19.44\%.
The WET baseline performed the weakest (0.4057), indicating that customized HTML parsers significantly outperform raw HTML extraction, as WET data, processed without parsing, retains more noise and lower-quality text.

\section{Filtering}

Effective filtering is the primary lever for reducing noise and irrelevant web material in large pretraining corpora.
The quality of the final dataset has a direct impact on downstream benchmark performance, so the pipeline is structured to maximize quality improvement relative to compute cost.
The key idea is to remove low-value text early, before applying any computationally expensive operations.
We place fast quality filters at the beginning of the pipeline.
These include lightweight heuristic-based screens such as Gopher Quality rules \cite{39}, Document quality filtering, and Line level filtering.
These filters quickly eliminate spam, boilerplate content \cite{6}, malformed pages, and toxic or extremely low-quality text.
Early filtering reduces the dataset size significantly, ensuring that only candidate documents of reasonable quality proceed to later stages.

Once the dataset is pruned to a cleaner subset, we apply repetition analysis.
Document-level repetition filters, such as GopherRepetition \cite{39}, detect structural redundancy, repeated lines, paragraph loops, and low-entropy n-gram patterns.
This step is computationally expensive, so performing it later avoids wasted computation on documents that would have been discarded anyway.
Repetition filtering prevents the model from learning repetitive or collapsed language patterns, stabilizes training distribution, and ensures more efficient token by the presence or absence of built-in stopword heuristics and isolates the impact of extraction quality itself.usage.
Global Deduplication removes near-duplicate documents across the corpus, while repetition filters target redundancy inside individual documents.
This two-level strategy ensures both diversity across documents and richness within documents.
The final outcome is a cleaner and more informative pretraining corpus that supports stronger model generalization and more stable benchmark performance.

The table below presents an overview of all filters applied in the pipeline.
For each filter, we report its threshold, the number of tokens removed, and the corresponding percentage out of the 3.6B input tokens.
This allows a clear comparison of how each filter contributes to overall data reduction.

\begin{table}[H]
\centering
\small
\caption{Sequential filter retention rates on Justext-parsed WARC files.}
\label{tab:filter_retention}

\renewcommand{\arraystretch}{1.15} 

\renewcommand{\tabularxcolumn}[1]{>{\centering\arraybackslash}m{#1}}

\begin{tabularx}{\textwidth}{| c | X | c | c |}
\hline
\textbf{Filters} & \textbf{Threshold} & \textbf{Removed Tokens} & \textbf{Removed Percentage(\%)} \\
\hline
URL filter & NIL & 258421645 & 7.17 \\ \hline
Language id filter & 0.65 & 144279806 & 4.00 \\ \hline
Duplicate line fraction & 0.3 & 243089688 & 6.74 \\ \hline
Duplicate line char & 0.2 & 224420961 & 6.22 \\ \hline
Duplicate para fraction & 0.3 & 215825629 & 5.99 \\ \hline
Duplicate para char & 0.2 & 207217936 & 5.75 \\ \hline
Top n-gram Char Thresholds & (2, 0.20), (3, 0.18), (4, 0.16) & 10889314 & 0.3 \\ \hline
Dup n-gram Char Thresholds & (5, 0.15), (6, 0.14), (7, 0.13), (8, 0.12), (9, 0.11), (10, 0.10) & 468693213 & 13.00 \\ \hline
Min doc words & $\ge$50 & 22749118 & 0.63 \\ \hline
Max doc words & $\le$ 100000 & 70319855 & 1.95 \\ \hline
Min avg word len & $\ge$3 & 723635 & 0.02 \\ \hline
Max avg word len & $\le$10 & 7562512 & 0.21 \\ \hline
Max symbol ratio & $\le$0.1 & 734300 & 0.02 \\ \hline
Max bullet ratio & $\le$0.9 & 2348123 & 0.01 \\ \hline
Max ellipsis ratio & $\le$0.3 & 45070091 & 1.25 \\ \hline
Min alpha ratio & $\ge$0.8 & 9402247 & 0.26 \\ \hline
Min stop words & $\ge$2 & 1669898 & 0.05 \\ \hline
Max non alphanum ratio & $\le$0.25 & 6092450 & 0.17 \\ \hline
Max urls ratio & $\le$0.2 & 1522522 & 0.04 \\ \hline
Max whitespace ratio & $\le$0.25 & 2959830 & 0.08 \\ \hline
Default line punctuation & $\ge$0.12 & 180338446 & 5.00 \\ \hline
Default Short Line & 0.67 & 129434806 & 3.59 \\ \hline
Char Duplicates & 0.01 & 830269122 & 23.03 \\ \hline
New Line Ratio & $\ge$0.3 & 835079753 & 23.16 \\
\hline
\end{tabularx}
\end{table}

\newpage
\section{Deduplication} \label{app:deduplication}
Deduplication is a critical component in constructing a high-quality pretraining corpus.
When two datasets contain an equal number of documents, the dataset with fewer duplicates provides greater informational diversity and leads to improved downstream benchmark performance \cite{40}.i
Repeated documents add no new knowledge, waste compute and tokens, and increase the risk of memorization, where models learn specific text fragments instead of generalizable patterns.
Effective deduplication thus ensures a richer, more efficient training signal and broader linguistic coverage.

To evaluate the impact of deduplication choices, we conducted four ablation experiments, each designed to analyze trade-offs between memory efficiency, strictness, and near-duplicate detection.
The configurations evaluated were:

\textbf{1. Bloom Filter (Both configuration)}

{\textbf{2. Bloom Filter (Old Both configuration)}}

{\textbf{3. Exact + MinHash deduplication}}

{\textbf{4. Exact + Substring + MinHash deduplication}}

All approaches used a consistent n-gram size of 13 tokens to ensure comparability.

\subsection{Bloom Filter-Based Deduplication}

The Bloom Filter \cite{13} is a probabilistic data structure widely used for efficient set membership testing, particularly suited to large-scale text deduplication.
It maintains a fixed-length bit array, initialized to zero. Each n-gram encountered is passed through multiple hash functions, producing several indices whose corresponding bits in the array are set to one.
When all bits corresponding to a new n-gram are already set, the n-gram is treated as a duplicate; otherwise, it is marked as unique.
This approach offers fast membership checks and excellent memory efficiency but introduces false positives due to possible bit collisions; distinct n-grams may hash to overlapping positions and be mistakenly treated as duplicates.
To mitigate this, the Bloom Filter parameters were tuned to an expected n-gram count of 3 billion, a false positive rate of $10 \times e^{-13}$, and 13-token n-grams, balancing accuracy with memory usage.

\textbf{We implemented two hierarchical configurations for large-scale corpus deduplication:}

\textbf{Both Configuration} 

Deduplication begins at the paragraph level. Each paragraph’s n-grams are checked against the Bloom Filter.
If the ratio of repeated n-grams exceeds a predefined threshold, the paragraph is considered duplicate.
When the proportion of duplicate paragraphs in a document exceeds a second threshold, the document itself is removed.
In this configuration, the Bloom Filter is not updated with n-grams from removed documents, ensuring that duplicates do not influence future comparisons.
\newpage
\textbf{Old Both Configuration}

This follows the same detection logic but updates the Bloom Filter with n-grams from non-duplicate portions of removed documents. This enables the filter to retain partially unique content for future comparisons, improving recall while maintaining precision.

\subsection{Hybrid Deduplication: Exact + MinHash and Exact + Substring + MinHash}

Beyond the hierarchical Bloom Filter configurations, we further examined hybrid deduplication approaches that combine exact, substring, and MinHash-based detection for greater coverage.

\textbf{Exact Deduplication}

Operates at the document level by comparing entire document hashes.
It efficiently removes identical documents but cannot detect partial or near-duplicate cases.

\textbf{Substring Deduplication} 

Extends this by examining 13-token substrings (n-grams).
If any substring in one document matches a substring in another, the document is flagged as duplicate \cite{12}.
This stricter approach captures near-verbatim repetition even when the surrounding text differs.
While computationally heavier, it enhances granularity and effectively removes localized repetition in web data.
These substring checks are complemented with MinHash, which provides near-duplicate detection for paraphrased or lightly altered content, offering a strong balance between precision and recall.

\textbf{MinHash Deduplication}

MinHash \cite{14} is a locality-sensitive hashing (LSH) \cite{17} technique that efficiently approximates the Jaccard similarity between sets and is widely used for large-scale near-duplicate detection.
In our experiments, each document was represented as a set of 13-token shingles, from which compact MinHash signatures were generated.
These signatures enable efficient comparison between documents without computing full pairwise similarity.
Documents whose estimated Jaccard similarity exceeded a fixed threshold were treated as near-duplicates, and only one version was retained.
MinHash effectively captured lightly paraphrased or reordered text, complementing the Bloom Filter’s efficiency in exact and hierarchical deduplication.
This MinHash-based deduplication was evaluated alongside Exact and Substring matching approaches.
Ultimately, we selected Bloom Filter–based deduplication for the final pipeline, as it achieved the best balance between low memory footprint, scalability, and consistent performance in repeated corpus construction.

\begin{figure}[H]
\centering
\includegraphics[width=\linewidth]{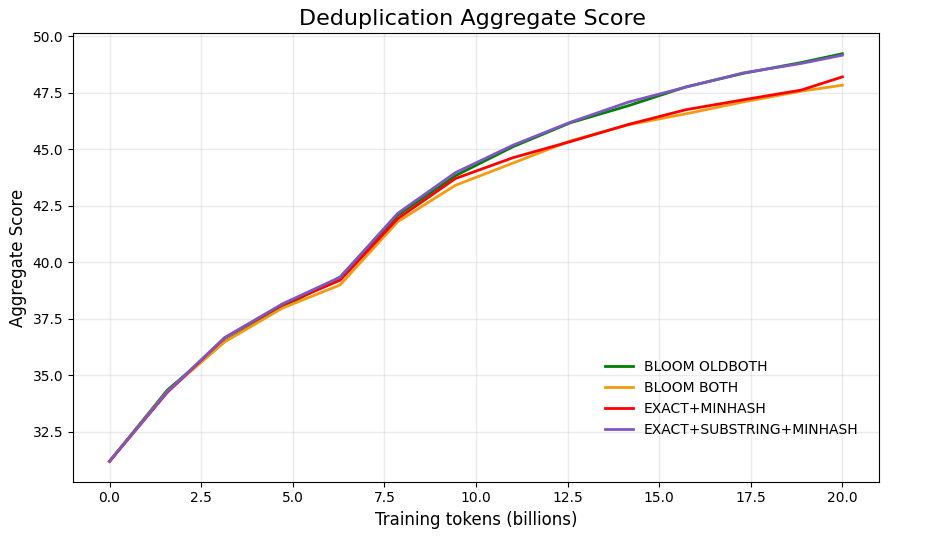}
\caption{\textbf{Deduplication Aggregate Score.} The results indicate that MinHash-based deduplication alone does not yield optimal performance; its effectiveness increases significantly when combined with a Bloom filter component. Incorporating Bloom filtering enhances the removal of residual near-duplicates, thereby improving overall model performance. Notably, the Bloom Filter (Old Both) configuration achieves performance comparable to the hybrid deduplication setup, demonstrating its efficiency as a balanced and reliable deduplication strategy.}
\label{fig:dedup_score}
\end{figure}

\section{Classifier} \label{app:classifier}
We conducted multiple ablation experiments using diverse data sources and model architectures to evaluate the impact of classifier design on final dataset quality.
The following section details each ablation setup, describing the specific data configurations, models, and evaluation methodologies employed.

\subsection{ FastText Classifier}

We employed a FastText-based classifier \cite{10} as the final semantic filtering stage in the pipeline.
FastText was chosen for its simplicity, efficiency, and scalability across large datasets.
It uses subword embeddings, which allows it to handle noisy or unseen tokens effectively while maintaining good performance.
Two different FastText setups were explored:
OpenHermes - ELI5 Configuration: In this setup, OpenHermes 2.5 \cite{63} and ELI5 explanations \cite{64} were used as positive examples representing high-quality and educational text.
Negative examples were randomly sampled from the pre-classification corpus to represent lower-quality content.
BETR-Style Configuration: Following a BETR-style \cite{57} approach, we selected the highest-scoring subset of our own dataset as positive examples and paired them with an equal-sized random subset from the remaining data as negatives.
This configuration aimed to internally leverage our own data distribution for classification.

\subsection{BETR (Benchmark Targeted Ranking)}

In the BETR \cite{57} approach, we began by selecting the train splits of established benchmark datasets available on Hugging Face.
In parallel, we prepared our own corpus of approximately 1M documents obtained after running the full pipeline up to the deduplication stage.
We then generated embeddings for both the benchmark datasets and our corpus using the all-MiniLM-L6-v2 \cite{46} embedding model.
Cosine similarity was computed between each document in our dataset and the benchmark embeddings.
Each document received multiple similarity scores, and the highest similarity score was assigned as its representative score.
We then applied a scoring threshold by selecting the top 10\% of documents.
This resulted in approximately 90K documents being labeled as high-quality.
For the low-quality class, we randomly sampled 90K documents from the remaining 90\% of the dataset.
This provided a balanced dataset of positive and negative samples for subsequent classifier training.
For FastText-based classification, we trained a supervised model using the official FastText library.
The model was optimized for efficiency and generalization on noisy web-scale data.
Training was performed with a learning rate of 0.1, embedding dimension of 100, and a context window size of 5.
We used 5 epochs with a minimum word count threshold of 1 and bi-gram features (word Ngrams = 2) to capture short-range dependencies.
The model employed the softmax loss function for multi-class classification and was initialized with a fixed random seed (42) to ensure reproducibility.
These settings provided a balance between speed, robustness, and classification accuracy for large-scale text filtering.

\subsection{DeBERTa Classifier}

For model-based classification, we used a DeBERTa-v3-base \cite{45} model with a linear classification head and a maximum sequence length of 512 tokens.
The classifier was trained on a dataset stratified into three quality categories:
High-Quality: Documents from Nemotron-CC HQ \cite{47}, Ultra-FineWeb \cite{55}, and FineWeb-Edu \cite{1}, which contain factually accurate and educational content.
Mid-Quality: Documents from DCLM \cite{2} and related sources, which are generally coherent but show higher noise and stylistic variability.
Low-Quality: Spam, scam, and low-retention portions of FineWeb \cite{1} and RefinedWeb \cite{3}, containing noisy or semantically weak content.
Training was performed using the AdamW optimizer with cosine learning rate scheduling (6\% warmup), a learning rate of $2 \times 10^{-5}$, batch sizes of 16/8 with gradient accumulation, and FP16 precision for efficient memory use \cite{66,67}.
The model was trained for 2 epochs. This DeBERTa-based classifier was then applied to label and filter large-scale web text by quality level.

\subsection{LLaMA-Score + BERT Classifier}
We also experimented with a two-step model-based classification approach. First, a LLaMA-3.3-70B \cite{61} model was used to assign a quality score between 0 and 5 to each document, reflecting its educational value, diversity, and textual clarity.
Documents that received scores of 3 and greater than 3 were selected as high-quality references. These high-quality samples were then used to fine-tune a BERT classifier \cite{62}, which was subsequently applied across the full dataset to categorize and filter documents according to learned quality patterns.
Below is the prompt we gave to the LLama-3.3-70B model for scoring the docs.

\begin{figure}[H]
\centering
\caption*{LLaMA Scoring Prompt}
\begin{lstlisting}
"Your task is to assess the quality of documents (scores 0-5), sampled from candidate corpora, in terms of their usefulness for improving performance on key educational and reasoning benchmarks: MMLU, ARC, BoolQ, CommonsenseQA, HellaSwag, OpenBookQA, PIQA, SocialIQA, and WinoGrande."

How do you judge the usefulness of a document? "While this is a subjective task, there are a few things to keep in mind:"

Apply the strict quality criteria and assign a score:

"High-Quality (4-5): Clear, unambiguous explanations with step-by-step reasoning that builds understanding across multiple steps; factually accurate with proper context; academic/educational tone with precise terminology; demonstrates diverse reasoning types (logical, causal, analogical, spatial); at least 300 characters of substantive instructional material with genuine learning value."

Medium-Quality (2-3): Basic factual coverage or simple Q&A without deep explanation; correct but shallow treatment; limited transfer to broader contexts; more reference-like than instructional.

Low-Quality (0-1): Marketing/promotional/spam; misinformation or ambiguities that hinder comprehension; casual chatter with little educational value; under 200 characters; predominantly non-English; simple lists without explanatory context; code snippets without instructional framing.

"Judge whether the document would help a model answer benchmark-style questions (multiple-choice science, commonsense, reading comprehension, world knowledge) by providing reasoning, not just facts."

"Prefer materials that show multi-step solution paths, worked examples, causal explanations, or comparative analyses over rote definitions or isolated facts."

"Down-rank documents with boilerplate, poor structure, or unclear context; exclude harmful content and sensitive personal data you would not want a model to generate."

Check length and language: high-quality items typically exceed 300 characters and are clearly written in English; very short items (<200 characters) are usually low quality.

"Avoid content that is primarily lists, tables, or raw code unless accompanied by explanations that teach underlying principles."

"When uncertain, err toward a lower score and write a brief one-sentence rationale for your decision."
\end{lstlisting}
\end{figure}

\begin{figure}[H]
\centering
\includegraphics[width=\linewidth]{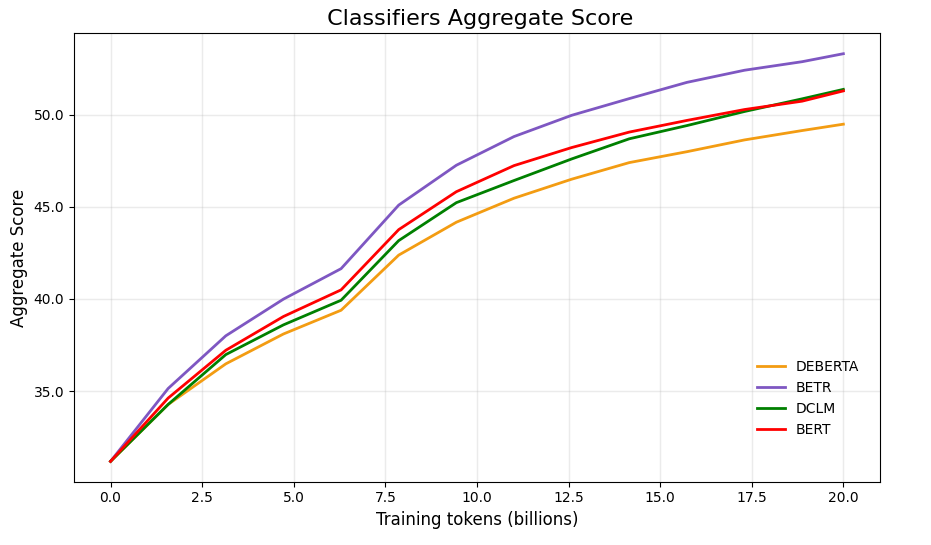}
\caption{\textbf{Classifier ablation comparison across four approaches.} BETR-based FastText classifier achieves highest aggregate accuracy (0.538), outperforming DeBERTa (0.4948), DCLM-bin fasttext classifier (0.5137), and LLaMA-Score+BERT (0.5128) methods.}
\label{fig:classifier_score}
\end{figure}

Across all classifier variants evaluated on the 1B model setup, we observed comparable performance trends with final aggregate accuracies converging around the 0.49–0.53 range on the core benchmarks.
Among them, the BETR-based classifier achieved the highest aggregate accuracy of 0.5330, followed by DCLM \cite{2} and BERT, while the DeBERTa classifier performed slightly lower but remained consistent.
These results indicate that classifier choice influences overall dataset quality and downstream model performance, though the gap between approaches is relatively narrow.
The findings suggest that lightweight and scalable classifiers such as FastText (BETR) can achieve competitive quality separation comparable to more complex transformer-based models like DeBERTa, making them effective for large-scale corpus filtering in practical pipeline deployments.
\newpage
\section{Domain and Format Distribution Analysis}
To analyze the topical and structural composition of our dataset, we employed the TopicClassifier, a 140M parameter model based on gte-base-en-v1.5 \cite{58}, fine-tuned on WebOrganizer/TopicAnnotations\cite{60}-Llama-3.1-8B (1M documents) and Llama-3.1-405B-FP8 \cite{59}(100K documents).
The classifier was used to infer both domain and format distributions, providing insights into the thematic balance and textual styles present in the corpus.

\begin{figure}[H]
\centering
\includegraphics[width=\linewidth]{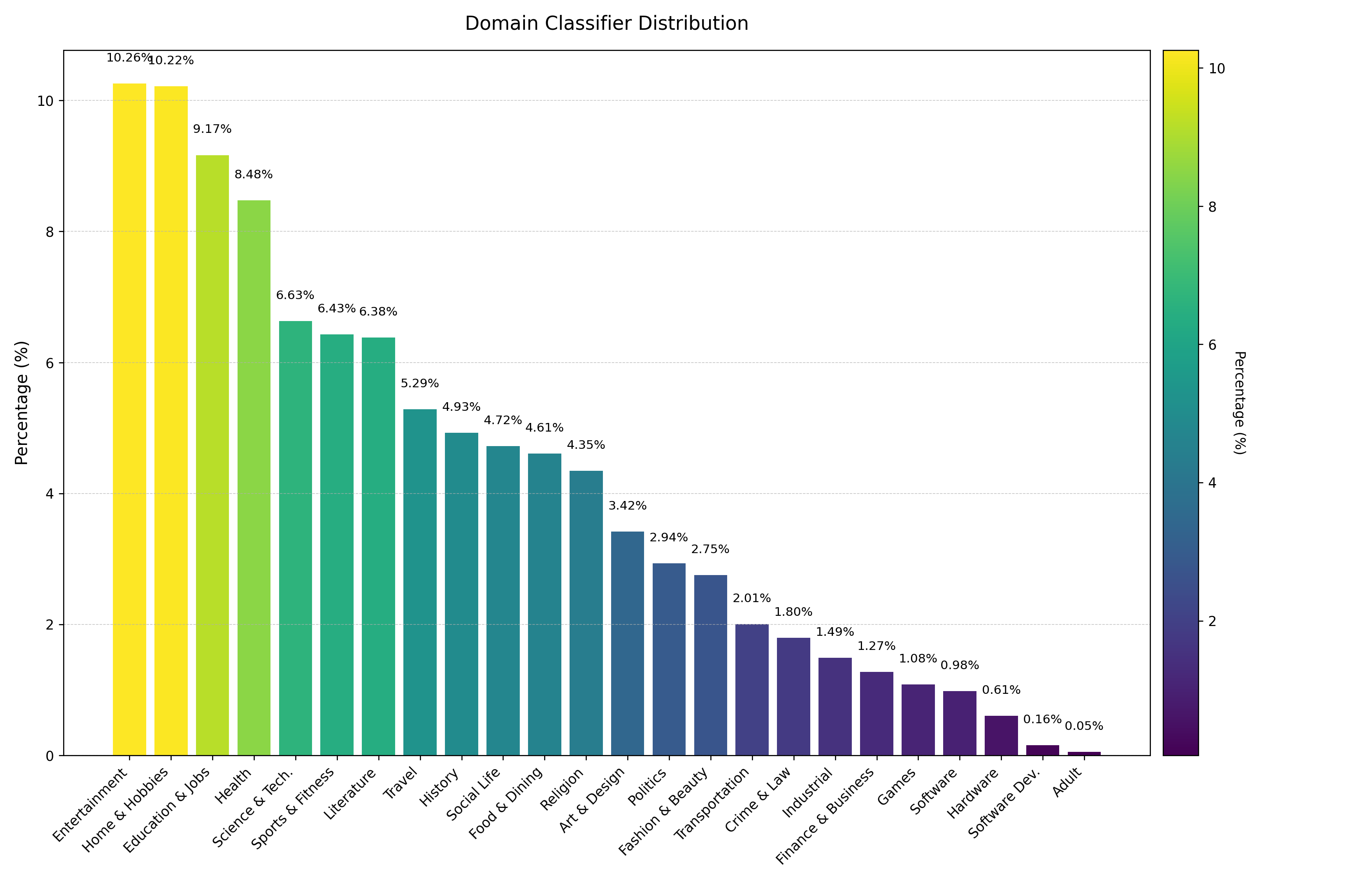}
\caption{\textbf{Domain distribution analysis across 17 categories.} Entertainment (10.26\%), Home \& Hobbies (10.22\%), and Education (9.17\%) comprise the largest domain categories in the filtered corpus.}
\label{fig:domain_dist}
\end{figure}

As shown in Figure F.1, the dataset exhibits a broad and balanced domain distribution across 17 categories.
The largest segments include Entertainment (10.26\%), Home \& Hobbies (10.22\%), Education \& Jobs (9.17\%), and Health (8.48\%), reflecting a diverse mix of informational and lifestyle-oriented content.
Technical and specialized areas such as Science \& Tech, History, and Politics contribute smaller but significant proportions, while domains like Hardware and Adult remain minimally represented, ensuring data safety and topical coverage.

\begin{figure}[H]
\centering
\includegraphics[width=\linewidth]{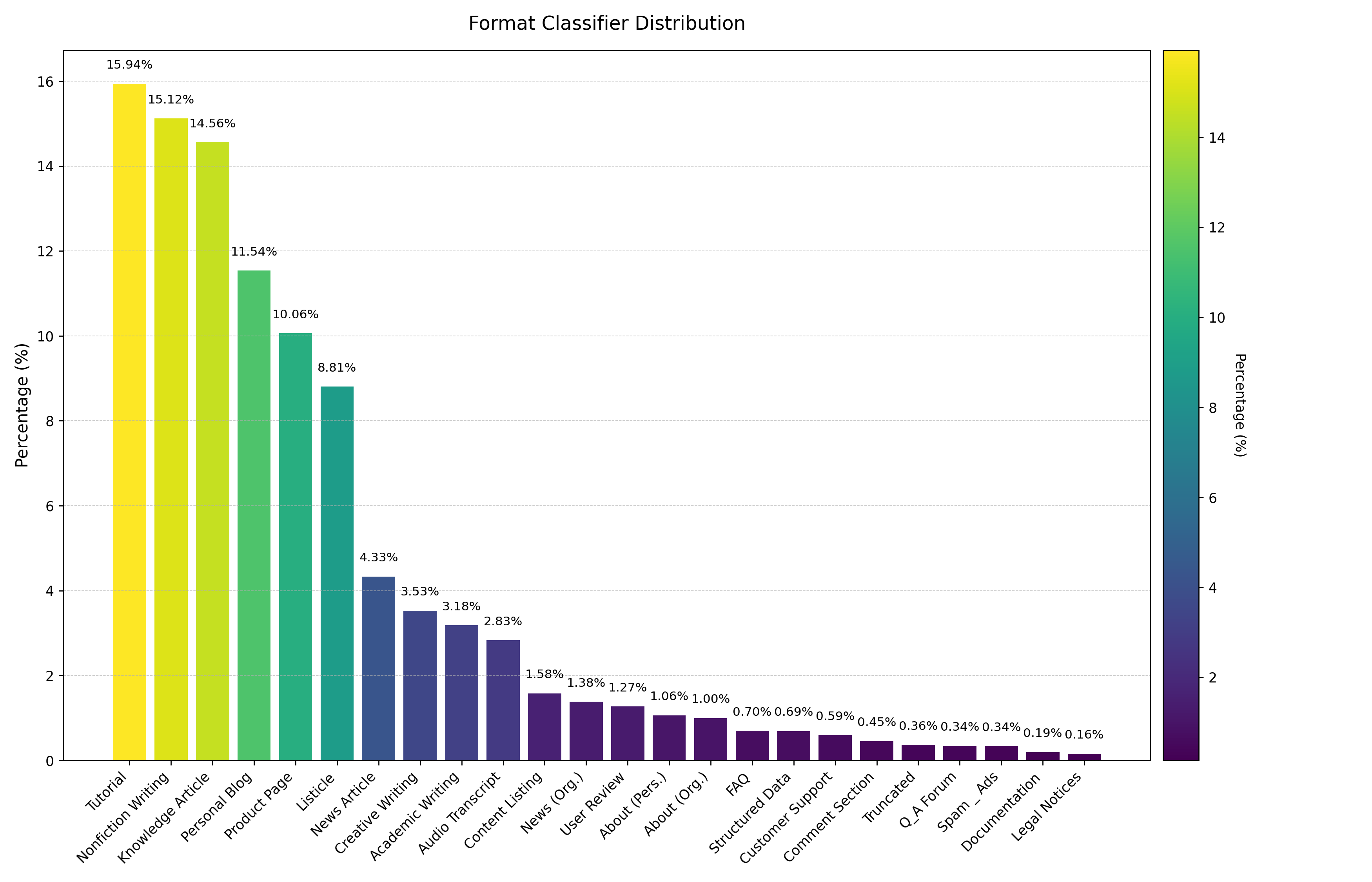}
\caption{\textbf{Format distribution breakdown across document types.} Tutorials (15.94\%), Nonfiction (15.12\%), and Knowledge Articles (14.56\%) represent the dominant content formats after pipeline processing.}
\label{fig:format_dist}
\end{figure}

The format distribution shown in Figure F.2 further demonstrates the structural diversity of the dataset.
Predominant formats include Tutorials (15.94\%), Nonfiction Writing (15.12\%), Knowledge Articles (14.56\%), and Personal Blogs (11.54\%), suggesting that a substantial portion of the corpus captures explanatory and narrative styles suited for general-purpose model training.
Other formats such as News Articles, Creative Writing, and Academic Writing appear in moderate proportions, while Q\&A Forums, Spam/Ads, and Legal Notices remain minimal, indicating effective filtering of noisy or low-quality content.
Overall, the distribution analysis confirms that the dataset maintains broad topical diversity and healthy structural variation, key properties for training robust and generalizable language models.

\section{Scaling-Law Based Data Selection} \label{app:scaling}
Selecting an optimal pretraining corpus cannot rely solely on small-scale (e.g., 1B-parameter) validation results, as dataset rankings frequently change when model capacity increases.
To address this scale-dependent instability, we incorporate a scaling-law–based evaluation protocol that predicts target-scale performance using only sub-target experiments.
We train a consistent size ladder of models and evaluate each dataset under two complementary approaches: (i) a single-scale baseline, where datasets are ranked using a fixed model size, and (ii) a multi-scale two-step scaling method.
The multi-scale method first fits a power-law relationship between compute and loss (loss vs compute) \cite{34, 56}, and then maps loss to downstream performance using a fixed-floor logistic function (accuracy vs loss) \cite{39}.
This framework allows us to extrapolate dataset performance to larger model scales without training full-scale models for every corpus.
The resulting predictions provide a compute-efficient and scale-aware mechanism for identifying the dataset most likely to yield superior performance at target model sizes.

\subsection{Methodology}
This section describes the two dataset evaluation strategies used in our study:
(i) a single-scale baseline, which ranks datasets using a fixed small model, and
(ii) a multi-scale extrapolation method that jointly models the compute-loss and loss-accuracy relationships to predict performance at larger model scales.

\subsubsection{Single-Scale Ranking Baseline}
In the single-scale setting, a model of fixed size (e.g., 150M, 400M, or 1B parameters) is trained independently on each dataset D under identical optimization hyperparameters and a matched compute budget.
For each run, we record either the downstream validation accuracy or a likelihood-based proxy signal derived from the model's outputs.
Datasets are then ranked according to these measured values.
This approach is computationally efficient because it requires only one training run per dataset.
However, prior scaling-law literature (including DataDecide) demonstrates that performance rankings at small scale do not reliably translate to larger models \cite{49}.
Many datasets exhibit scale-dependent crossovers, where a dataset that appears superior at 150M becomes inferior at 1B due to differing loss-reduction exponents.
These empirically observed inversions motivate the need for a prediction mechanism that incorporates information from multiple scales.

\subsubsection{Multi-Scale Extrapolation (Two-Step Scaling Approach)}
To obtain scale-aware predictions of dataset performance at larger target sizes, we employ a two-step scaling-law framework.
This method learns two empirical relationships from sub-target models (e.g. $\le$1B parameters):

1. Compute $\rightarrow$ Loss (scaling law), and

2. Loss $\rightarrow$ Accuracy (logistic mapping).

By combining these two components, we can predict accuracy at arbitrary target compute budgets without training full-scale models for every dataset.

\vspace{1em}
\noindent \textbf{Compute-Loss Modeling:} For each dataset D, let C denote the compute used by models in the size ladder.
We model the final validation loss as a constrained power-law:
\begin{equation}
L(C) = A C^{-\alpha} + E
\end{equation}

where $A>0$ is the offset term, $\alpha \in [0.05,0.60]$ is the scaling exponent governing how quickly the loss decreases with additional compute, and $E \ge 0$ represents the irreducible loss floor.
These constraints are consistent with established neural scaling laws and ensure physically meaningful extrapolations.
Only sub-target model sizes are used for fitting (A,$\alpha$,E), preventing leakage of target-scale information and isolating the dataset-specific scaling trajectory.
When proxy metrics (e.g., normalized probability of the correct answer) are used instead of true log-likelihood, we apply a monotonic transformation such as negative log or reciprocal to map them onto a loss-compatible domain.

\begin{figure}[H]
\centering
\includegraphics[width=\linewidth]{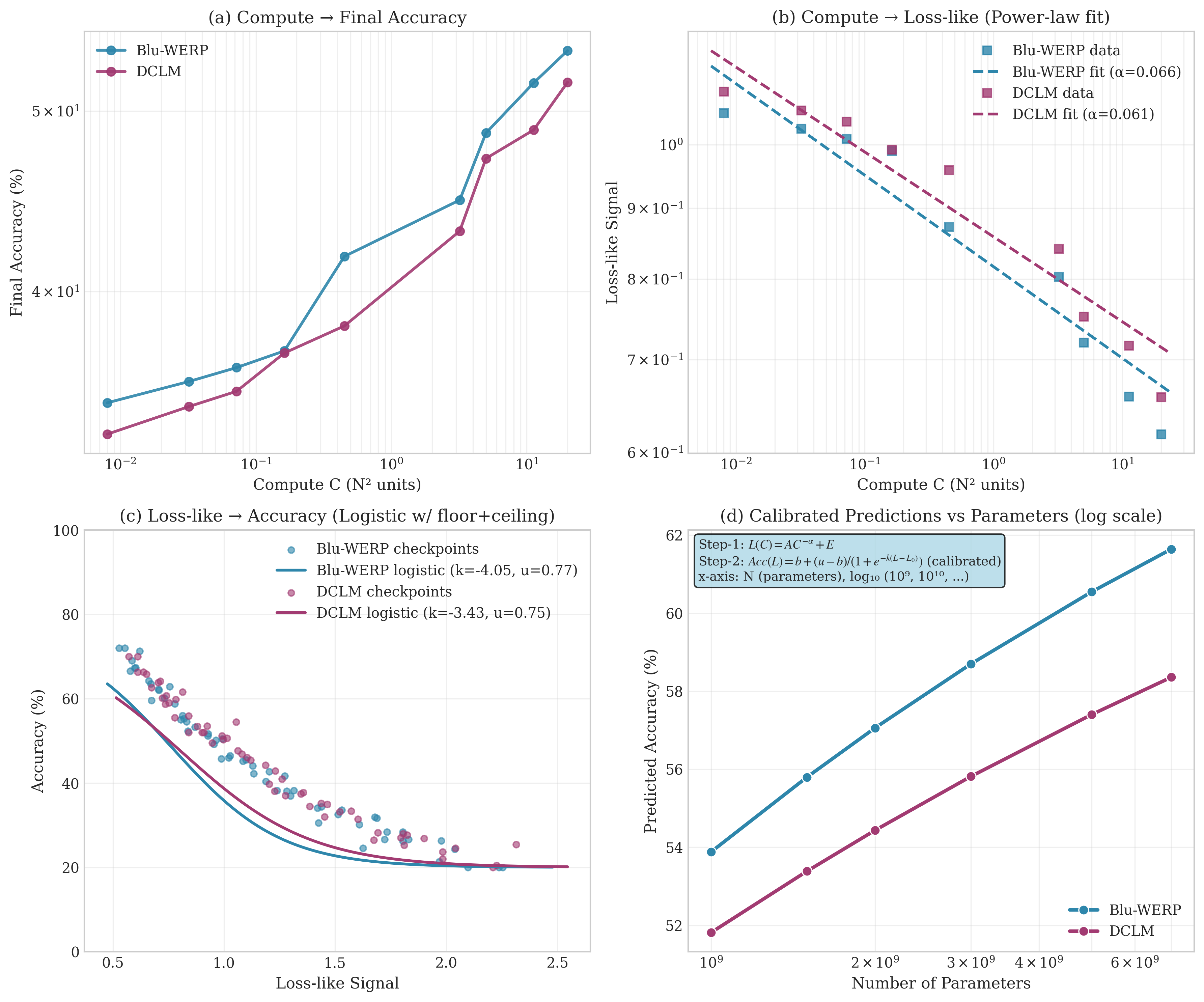}
\caption{\textbf{Multi-Scale Extrapolation Results.} Compute-Loss Modeling and Loss-Accuracy Mapping}
\label{fig:extrapolation}
\end{figure}

\vspace{1em}
\noindent \textbf{Loss–Accuracy Mapping:} To relate predicted validation loss to downstream evaluation accuracy, we fit a fixed-floor logistic function of the form:

\begin{equation}
\text{Acc}(L) = \frac{1-b}{1 + \exp(-k(L-L_0))} + b
\end{equation}

where b is the chance-level accuracy (e.g., b=0.25 for 4-way multiple choice), k governs the slope of the accuracy curve, and $L_0$ is the inflection point.
Only k and $L_0$ are optimized, while b is fixed to preserve the lower bound.
Validation accuracy is often noisy and discrete, whereas loss provides a continuous and more stable signal \cite{39}.
Therefore, the logistic function is fit using many intermediate checkpoints, not only final accuracies.
This produces a smooth mapping that captures how uncertainty reduction (loss) translates to downstream performance.
If a reliable accuracy measurement is available at a reference model size (e.g., 1B parameters), we optionally incorporate a calibration constraint by assigning higher weight to the corresponding loss–accuracy pair during fitting.
Once the mapping parameters (k,$L_0$) are obtained, the predicted accuracy at the target compute value is computed:

\begin{equation}
\text{Acc}(C_{\text{target}}) = \text{Acc}(L(C_{\text{target}}))
\end{equation}

\subsection{Experimental-Setup (Prediction)}
Prediction quality is measured using:
Pairwise Decision Accuracy - fraction of dataset pairs for which the method correctly predicts the better dataset at the target scale.
Rank Correlation - Spearman and Kendall between predicted and empirical rankings.
Mean Absolute Prediction Error (MAPE) - difference between predicted and empirical accuracies at 1B (when available).
Compute Cost (FLOPs) - total compute required to generate predictions, restricted to $\le$1B models.
Models were trained across a scale range of 20M to 1B parameters, following a compute-optimal Chinchilla-style regime \cite{56}, where the number of training tokens is set proportional to model size.
Specifically, each model was trained with 20× tokens per parameter, ensuring consistent scaling behaviour and comparable compute allocation across all model sizes.

\begin{figure}[H]
\centering
\includegraphics[width=\linewidth]{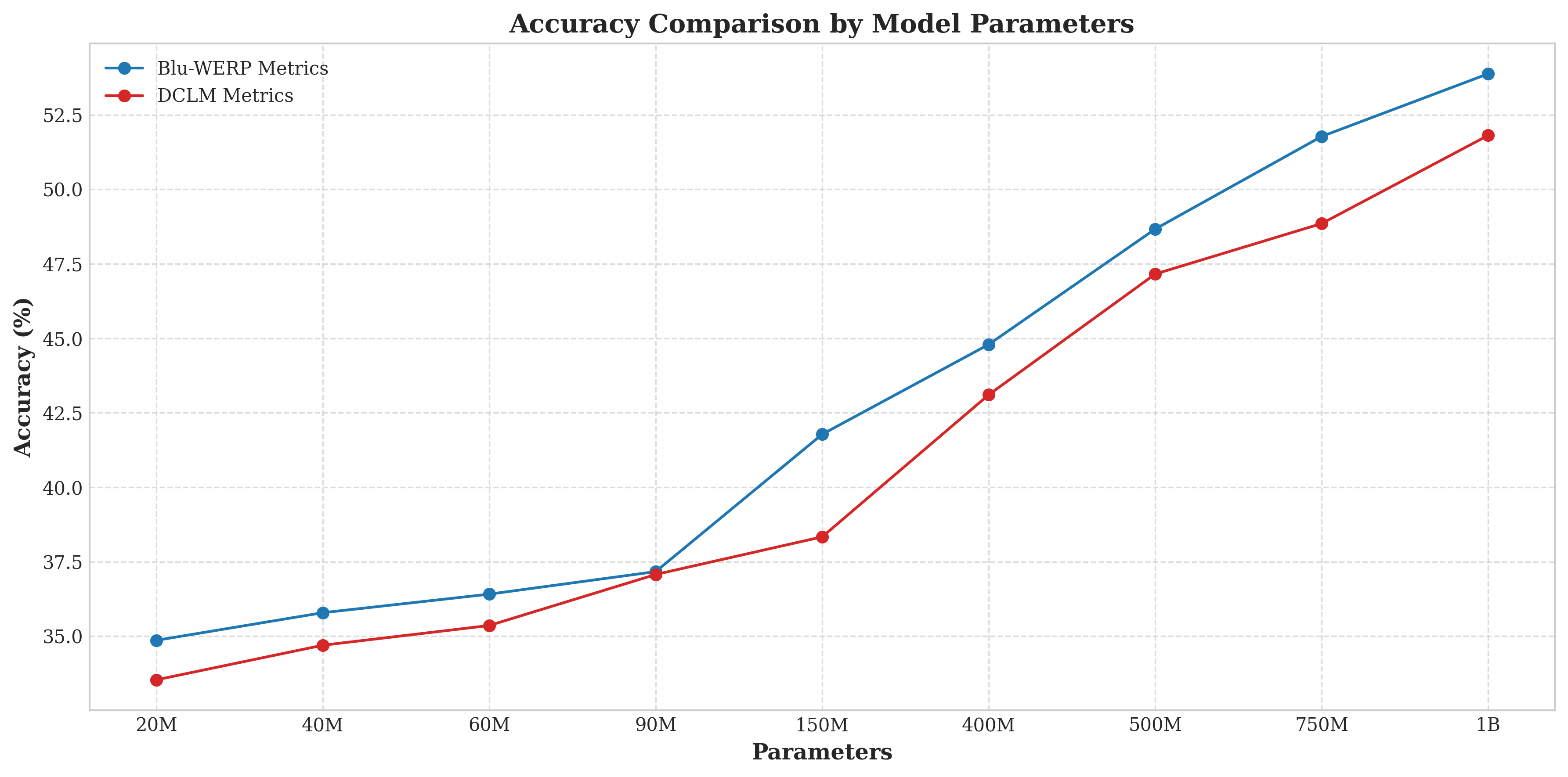}
\caption{Evaluation metrics for models scaled from 20M to 1B parameters on Blu-WERP and DCLM.}
\label{fig:eval_metrics}
\end{figure}

Using the observed results from sub-billion-scale models, the multi-scale fitting procedure enables accurate prediction of downstream performance for model sizes that were not explicitly trained.
We evaluate this two-step scaling methodology by forecasting accuracy at target scales exceeding the largest trained model (1B parameters).
Tables \ref{tab:pred_1_5b} and \ref{tab:pred_1_75b} summarize predicted accuracies at 1.5B and 1.75B parameters under both uncalibrated and calibrated variants of the method.
Across all settings, Blu-WERP consistently achieves higher projected performance compared to DCLM \cite{2}, indicating more favorable scaling behaviour.

\begin{table}[H]
\centering
\caption{Predicted Accuracy at 1.50B Parameters}
\label{tab:pred_1_5b}
\begin{tabular}{lcc}
\toprule
\textbf{Dataset} & \textbf{Pred. (uncal.)} & \textbf{Pred. (cal.)} \\
\midrule
Blu-WERP & 66.16\% & 56.38\% \\
DCLM & 64.66\% & 54.05\% \\
\bottomrule
\end{tabular}
\end{table}

Across both target sizes, the calibrated predictions show a consistent 2.3–2.5\% advantage for Blu-WERP, confirming stronger scaling behaviour.

\begin{table}[H]
\centering
\caption{Predicted Accuracy at 1.75B Parameters}
\label{tab:pred_1_75b}
\begin{tabular}{lcc}
\toprule
\textbf{Dataset} & \textbf{Pred. (uncal.)} & \textbf{Pred. (cal.)} \\
\midrule
Blu-WERP & 66.61\% & 57.06\% \\
DCLM & 65.09\% & 54.71\% \\
\bottomrule
\end{tabular}
\end{table}

\begin{figure}[H]
\centering
\includegraphics[width=\linewidth]{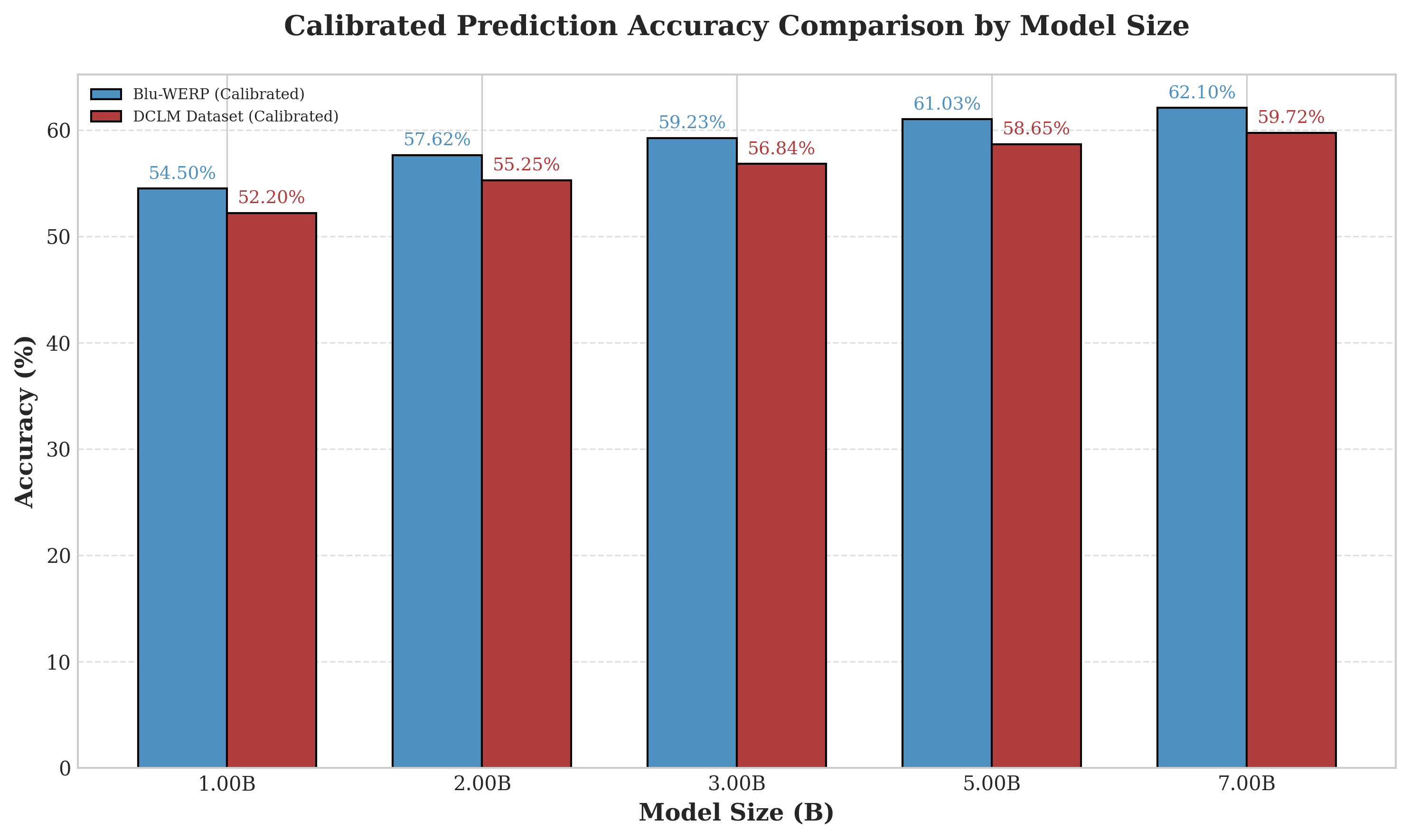}
\caption{Predicted performance for Blu-WERP vs. DCLM from 1B to 7B using multi-scale extrapolation.}
\label{fig:pred_perf}
\end{figure}

To evaluate the robustness of long-range extrapolation, we extend the predicted performance curve from 1B up to 7B parameters using the previously fitted scaling relationships.
Figure \ref{fig:pred_perf} reports the calibrated accuracy forecasts across these larger model sizes.
The performance margin between the two datasets increases steadily with scale, demonstrating that our dataset follows a more favourable compute–loss trajectory and translates this advantage into higher downstream accuracy as model capacity grows.

Comparative Interpretation:
Blu-WERP dataset: $\alpha \approx 0.066$ | DCLM $\alpha \approx 0.061 \rightarrow$ Higher $\alpha$ implies faster loss improvement with compute \cite{34, 56}.
Irreducible Loss (E) Blu-WERP dataset: extremely small ($\approx 0$) DCLM \cite{2}: noticeably higher for some fits $\rightarrow$ Indicates lower noise floor in Blu-WERP \cite{34}.
Logistic Mapping (k, $L_0$) Our logistic mapping places the inflection point at a lower loss, meaning the same loss corresponds to higher downstream accuracy \cite{39}.
\newpage
\subsection{Limitations}
The predictive pipeline presented here is subject to several important constraints.
First, all estimates beyond 1B parameters are extrapolations from sub-billion-scale models, and accuracy may degrade when extending more than 2–3× beyond the fitted region \cite{34, 56, 49}.
These predictions inherently depend on stable loss trajectories and consistent hyperparameters;
deviations in optimization behaviour or data quality can introduce additional variance.
Certain datasets exhibit non-monotonic scaling or rank reversals across model sizes, which can reduce the reliability of both single-scale and multi-scale estimates \cite{2, 49}.
Moreover, the logistic loss-accuracy mapping assumes a smooth saturation pattern that may not generalize to tasks with irregular accuracy dynamics or strong transfer mismatches \cite{39}.
Consequently, results at larger scales should be interpreted as informed projections rather than empirical measurements.

\end{document}